\documentclass[10pt,twocolumn,letterpaper]{article}

\usepackage{wacv}
\usepackage{times}
\usepackage{epsfig}
\usepackage{graphicx}
\usepackage{amsmath}
\usepackage{amssymb}
\usepackage{color}
\usepackage{caption}
\usepackage{tabularx}
\usepackage{float}
\usepackage{booktabs}
\usepackage[bookmarks=false]{hyperref}

\wacvfinalcopy 

\def\wacvPaperID{1246} 

\let\oldtwocolumn\twocolumn
\renewcommand\twocolumn[1][]{
    \oldtwocolumn[{#1}{
    \begin{center}
          \includegraphics[width=\textwidth]{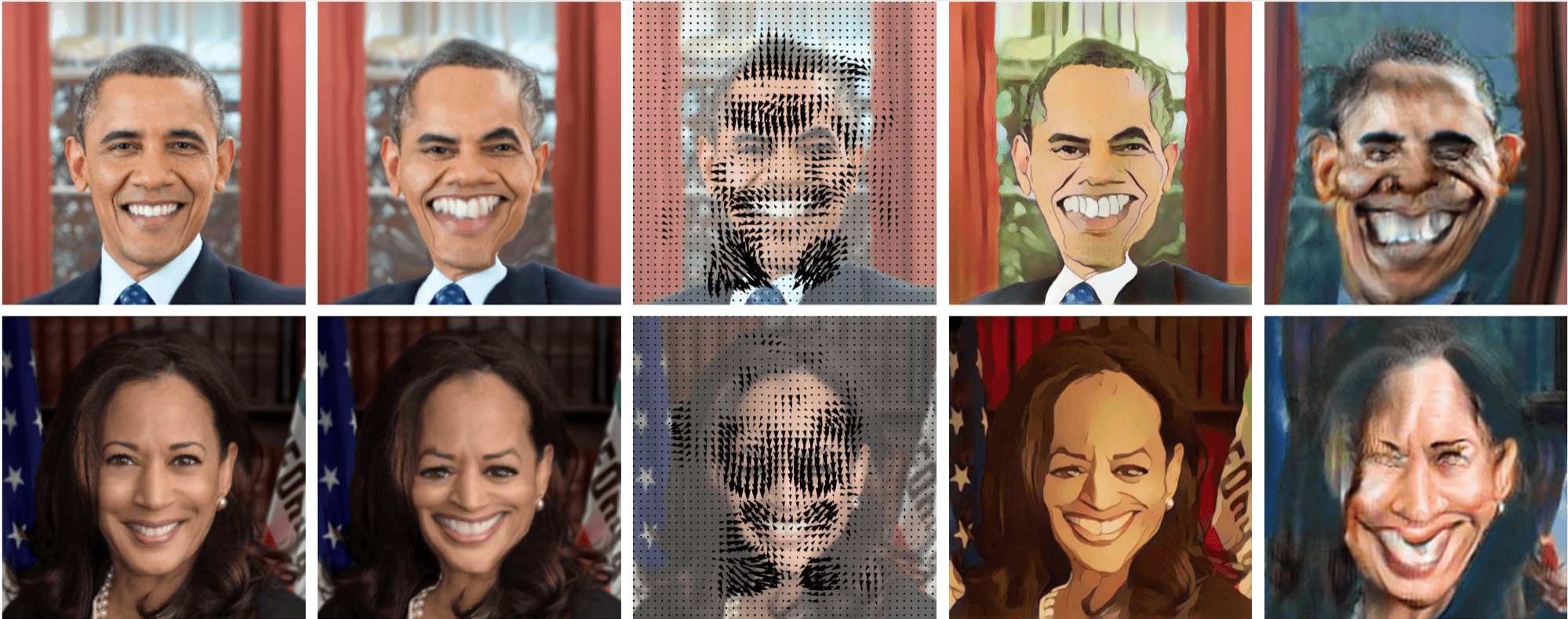}
          \centering
          \setlength{\tabcolsep}{0.2em}
          \begin{tabularx}{\textwidth}{*{5}{>{\centering\arraybackslash}X}@{}}
                (a) Input Photo & (b) Cartoon (Ours) & (c) Warping Field (Ours) & (d) Caricature (Ours + CartoonGAN \cite{ChenCartoonGAN2018}) & (e) Caricature (WarpGAN \cite{shi2019warpgan})
          \end{tabularx}
           
          \captionof{figure}{Example images from our test set (a), exaggerated cartoons (b) and overlaid warping fields (c) generated by our model (AutoToon), our model's cartoons stylized with CartoonGAN \cite{ChenCartoonGAN2018} to create caricatures (d), as compared to WarpGAN \cite{shi2019warpgan} caricatures (e).}
          \label{fig:teaser}
    \end{center}
    }]
}

\ifwacvfinal\fi
\begin{document}
\captionsetup{font=small}

\title{AutoToon: Automatic Geometric Warping for Face Cartoon Generation}

\author{Julia Gong \\
Stanford University\\
{\tt\small jxgong@stanford.edu}
\and
Yannick Hold-Geoffroy \\
Adobe Research\\
{\tt\small holdgeof@adobe.com}
\and
Jingwan Lu \\
Adobe Research\\
{\tt\small jlu@adobe.com}
}

\maketitle
\ifwacvfinal\thispagestyle{empty}\fi

\begin{abstract}
Caricature, a type of exaggerated artistic portrait, amplifies the distinctive, yet nuanced traits of human faces. This task is typically left to artists, as it has proven difficult to capture subjects' unique characteristics well using automated methods. Recent development of deep end-to-end methods has achieved promising results in capturing style and higher-level exaggerations. However, a key part of caricatures, face warping, has remained challenging for these systems. In this work, we propose AutoToon, the first supervised deep learning method that yields high-quality warps for the warping component of caricatures. Completely disentangled from style, it can be paired with any stylization method to create diverse caricatures. In contrast to prior art, we leverage an SENet and spatial transformer module and train directly on artist warping fields, applying losses both prior to and after warping. As shown by our user studies, we achieve appealing exaggerations that amplify distinguishing features of the face while preserving facial detail.
\end{abstract}

\section{Introduction}
Every human face is slightly different. While most people can identify faces familiar to them, it requires the more trained eye of a caricature artist to pick up on the most distinctive features that characterize an individual's face. In fact, caricature is a specific form of portraiture in which artists exaggerate the most visually salient characteristics of their subjects that distinguish them from others. Amplifying these defining features lets artists create more distilled portrayals of their subjects, and studies have shown that this skillful exaggeration can allow viewers to identify a subject's identity more easily from a caricature than from a normal photograph \cite{Rhodes1987IdentificationAR}.

With the rise of applying computer vision techniques to tackle creative tasks, an interesting problem that has emerged is automatic caricature generation. Similar to how an artist might approach caricatures, the computer vision analogy to caricature generation can be decomposed into two steps: 1) applying a geometric warp to the face that exaggerates salient features, and 2) stylizing the warped image for an artistic effect. The complete disentanglement of these two steps allows them to be independently learned and applied, leading to greater flexibility and higher quality of generated caricatures.

Early work in caricature generation mostly relied on rules-based methods \cite{Akleman1997Cari, Akleman2000Cari, gooch2004human, le2011cari, liang2002cari, mo2004cari}. More recently, with the rise of deep learning for artistic tasks such as sketch synthesis, image-to-image translation, and style transfer \cite{gatys2016image, isola2017image, Wu2019TransGaGaGU, ijcai2018sketchsynthesis, zhu2017cyclegan}, caricature generation has been re-introduced as an image-to-image translation problem first by Cao et al.~\cite{cao2018cari} and then Shi et al.~\cite{shi2019warpgan}. While these systems do achieve geometric exaggeration and artistic stylization, the exaggerations still have room for improvement. They often either do not precisely target the most salient facial features due to the constrained set of warping handles, or the warping is not disentangled completely from the artistic stylization, resulting in weaker standalone warps and less flexibility for combining different warps and styles.

In comparing the difficulty of these two stages of caricature generation, it is noteworthy that the computer vision community has seen much progress in general image stylization and style transfer in recent years, such as~\cite{gatys2016image,johnson2016perceptual}. However, effective geometric warping, especially applied to faces, has more room for improvement. In fact, there is less room for error in pure geometric warping; not only are our eyes highly attuned to faces \cite{TsaoFace2008}, but viewers are also more sensitive to the quality of unstylized, warped faces than that of stylized caricatures, since the resulting images are photorealistic. Thus, in this work, recognizing there are numerous high-quality methods that can perform stylization in the caricature generation pipeline, we focus on the more difficult stage: geometric warping of distinguishing characteristics to create a high-quality, warped version of the original photograph, the result of which we term a \textit{cartoon}.

Specifically, we aim to create an automated, end-to-end pipeline (AutoToon) that geometrically warps images of faces to generate cartoons, which are then used to create caricatures via existing stylization techniques. Our model learns a smooth warping field of pixel displacements that is applied to the input image, which can be scaled in magnitude to increase the exaggeration. By virtue of learning a warping field rather than performing image-to-image translation, our model preserves facial details more effectively and generates higher quality images for a given portrait.

Finally, to accompany our model, we also introduce the AutoToon dataset, a paired dataset of human facial portrait photos and their corresponding geometrically warped cartoons by trained artists. We hope that proposing a model for higher-quality face warping will accelerate the progress in creating end-to-end systems for caricature generation and other face-related cartoonization tasks.

Qualitative evaluation via user studies and artist appraisal of cartoons produced by AutoToon show that the generated cartoons from our approach exaggerate facial features more effectively than state-of-the-art warping methods. A summary of our contributions is as follows:
\begin{itemize}
\setlength\itemsep{0em}
    \item To our knowledge, AutoToon is the first supervised deep learning face cartoon generation model. It
    \begin{itemize}
    \setlength\itemsep{0em}
        \item automatically exaggerates salient facial features well in a caricature-like manner and can be scaled to control warping extent,
        \item is completely disentangled from stylization, and thus can be paired with any stylization method,
        \item is trained on less data, and preserves image details more effectively than previous methods.
    \end{itemize}
    \item A paired dataset, the AutoToon dataset, which also includes artist warping fields for photorealistic facial exaggeration and cartoon generation.
\end{itemize}
\section{Related Work}

Human faces have received a lot of attention in the literature over the years. Many approaches were developed to either model~\cite{sirovich1987low}, interact~\cite{han2018caricatureshop} or generate them~\cite{karras2019style}.  In this section, we review the work relevant to caricature generation and face warping. 

\subsection{Learned Warping}

Multiple works have learned and applied spatial transforms on images. First, parametric approaches such as the spatial transformer~\cite{jaderberg2015spatial} have been proposed to estimate global transform parameters. Flow-based approaches such as~\cite{revaud2015epicflow} further this idea by learning a dense deformation field over the whole image. DeepWarp~\cite{ganin2016deepwarp} proposes to apply this to gaze manipulation. Recently, Zhao et al.~\cite{zhao2019learning} uses this dense flow estimation to remove geometric distortion from close-range portrait images.
Cole et al.~\cite{cole2017synthesizing} also warp portrait images using spline interpolation on pre-detected landmarks while preserving identity.
Similar to our loss functions, Zhang et al.~\cite{zhang2014} use smoothness, local, and global alignment terms for parallax-tolerant image stitching.
Given the efficacy of flow estimation in these related application domains, our work on AutoToon aims to integrate this work with caricature generation by using dense flow estimation and the differentiable warping module from \cite{jaderberg2015spatial} to predict warping fields for generating cartoons.

\subsection{Caricature Generation}

One goal of caricature generation is to detect and amplify the unique features of a given face. Traditional techniques typically approached this by amplifying the difference from the mean, either by explicitly detecting and warping landmarks~\cite{brennan1985dynamic,gooch2004human,liao2004automatic,mo2004improved} or using data-driven methods to estimate unique face features~\cite{liu2006mapping,yang2016example,zhang2016data}. Early work largely relied on rules-based methods \cite{Akleman1997Cari, Akleman2000Cari, le2011cari, liang2002cari}, which limited caricature diversity. 
More recently, deep learning techniques have also been applied. For instance, Wu et al.~\cite{wu2018alive} model the subject face in 3D to improve how natural the caricature expression looks using a neural network.

Newer techniques for caricature generation are data-driven. There exist some readily available datasets of annotated caricatures, such as WebCaricature~\cite{HuoBMVC2018WebCaricature}, comprised of 6042 caricatures and 5974 photographs from 252 different identities. Despite these efforts, the limited amount of data available is still a major challenge. Thus, most of the work on this topic has taken inspiration from the recent generative image-to-image translation literature trained on unpaired images~\cite{choi2018stargan,huang2018multimodal,zhu2017cyclegan} and focuses on learning from unpaired portraits and caricatures~\cite{cao2018cari,wu2019attribute,Wu2019TransGaGaGU}.
Wu et al.~\cite{wu2019attribute} proposed to improve this image-to-image translation approach~\cite{isola2017image} by adding a geometric motion module.

Closer to our work, the first deep learning approach to caricature generation, CariGAN \cite{cao2018cari}, proposed to train a Generative Adversarial Network (GAN) using unpaired images to learn the image-to-caricature translation.
Building on previous work on style transfer and learned warping, Shi et al.~\cite{shi2019warpgan} then proposed a method that uses the GAN framework to jointly train style and warping end-to-end. However, while unpaired learning can leverage more data, they introduce highly varied exaggerations from artists with divergent styles, even for the same subject, making learning consistent exaggerations difficult. They also frequently have varying scales, poses, and low input-output correspondence, resulting in models learning very high-level features that may not be the most specific distinguishing features of a given face. The exaggerations learned by these models are relatively coarse as well due to the use of sparse warping points. Thus, in our work, we instead take a \textit{paired} supervised learning approach based on the work of two artists to balance this tradeoff, electing to learn specific artist styles well rather than an average of all styles. We also leverage the differentiable warping module from \cite{jaderberg2015spatial} to generate denser warping fields for more detailed exaggerations.

In contrast to previous work, we focus purely on the warping step of caricature generation to create high-quality warps while completely disentangling geometry and style.
\section{Problem Formulation and Warping Model}

In caricature generation, the task is to generate an exaggerated and stylized caricature for a given input portrait. Our new method, AutoToon, tackles the exaggeration portion of this pipeline. Given a normalized RGB portrait image $X_{in} \in \mathbb{R}^{H \times W \times 3}$, our task is to apply an artist-like facial exaggeration to $X_{in}$ to generate a cartoon image $\hat{X}_{toon}$. $\hat{X}_{toon}$ is then the input image to any stylization network to complete the caricature generation task.

\subsection{Warping and Linear Interpolation}
To discuss our method, we first need to formalize our definition of warping fields and grid sampling, which are key to our approach.

To perform the facial exaggeration for $X_{in}$, our network learns a flow field, which we call a warping field. The learned warping field $\hat{F} \in \mathbb{R}^{H \times W \times 2}$ is applied to $X_{in}$ to obtain $\hat{X}_{toon}$. The first channel of dimension $W \times H$ is a grid of scalar values representing the per-pixel displacement of $X_{in}$ in the $x$ direction, while the second channel encodes the same for the $y$ direction.

\begin{figure*}[!ht]
\begin{center}
\vspace{-1em}
\includegraphics[width=0.907\linewidth]{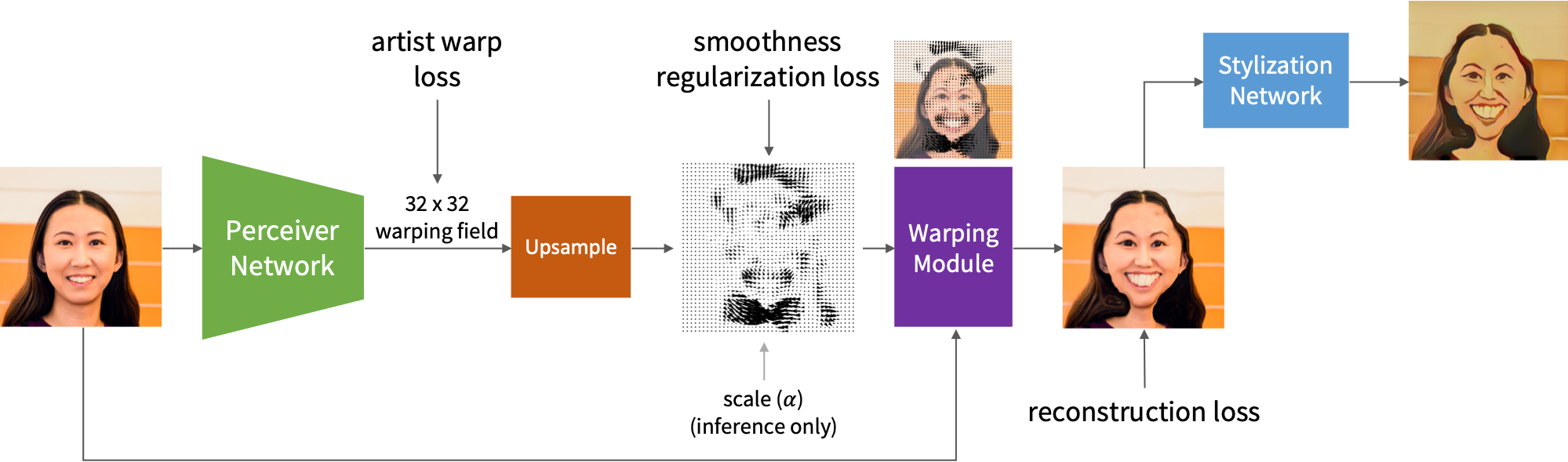}
\end{center}
  \caption{AutoToon model architecture and training losses. Given an input image, the Perceiver Network generates a $32 \times 32$ warping field. The warping field is upsampled via bilinear interpolation to obtain pixel-wise displacements, which is used to warp the input image into the resulting cartoon. The cartoon can then be stylized using any desired stylization network, such as CartoonGAN \cite{ChenCartoonGAN2018}, used here. At inference time, a scaling factor $\alpha$ can be applied to the warping field to manipulate warping intensity.}
\label{fig:model}
\end{figure*}

To perform exaggeration, this warping field is applied to $X_{in}$ via the differentiable warping module taken from Spatial Transformer Networks \cite{jaderberg2015spatial}. The module performs bilinear interpolation to displace the pixels of $X_{in}$ according to the learned displacements $\hat{F}$, or $\textit{Warp}(X_{in}, \hat{F})$. We call \textit{Warp} the Warping Module, as shown in Figure \ref{fig:model}.

\section{Proposed Method}
\subsection{Dataset}
\begin{figure}[t]
\begin{center}
\centering
\setlength{\tabcolsep}{0.1em}
\begin{tabular}{cccc}
    Input & Artist & Input & Artist \\
    \includegraphics[width=0.1155\textwidth]{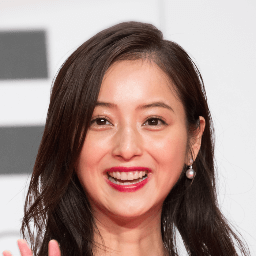} & \includegraphics[width=0.1155\textwidth]{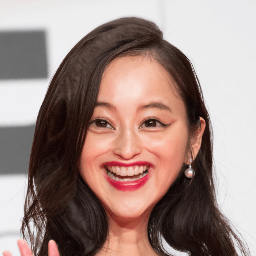} & \includegraphics[width=0.1155\textwidth]{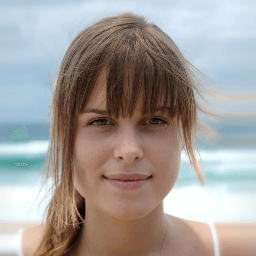} & \includegraphics[width=0.1155\textwidth]{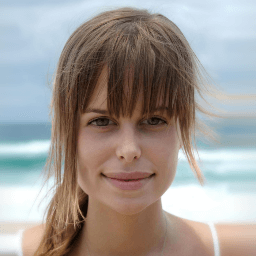} \\
    \includegraphics[width=0.1155\textwidth]{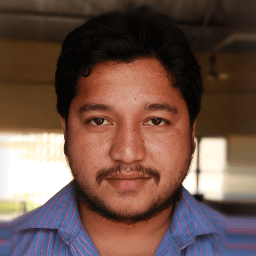} & \includegraphics[width=0.1155\textwidth]{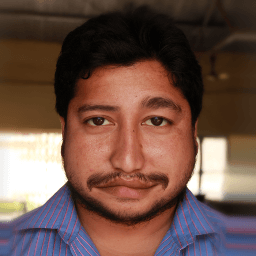} & \includegraphics[width=0.1155\textwidth]{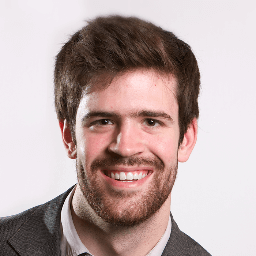} & \includegraphics[width=0.1155\textwidth]{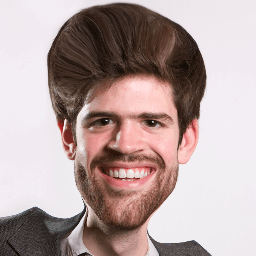} \\
\end{tabular}
\end{center}
\vspace{-1.5em}
  \caption{Four example pairs of input images and artist-warped cartoons from the training dataset. Photos by \href{https://flic.kr/p/S2593T}{Dick Thomas Johnson}, \href{https://flic.kr/p/5Td3Jr}{Shannon Luk}, \href{https://flic.kr/p/cEgdaL}{Possible}, and \href{https://flic.kr/p/a3MGD7}{Chuck Grimmett}; modified.}
\label{fig:dataset}
\end{figure}

101 portrait images of frontal-facing people (non-celebrities) were collected from Flickr. The people selected covered a broad range of age groups, sexes, races, and face shapes. These images were then warped via Adobe Photoshop by two caricature artists with similar styles to generate the ground-truth artist cartoons. This paired dataset of 101 images ($X_{in}, X_{toon}$) was split into 90 training and 11 validation images. The test set, without ground truth labels, was collected from various subjects and public sources. Sample images from the training set are shown in Figure \ref{fig:dataset}.

An additional component of the dataset that we provide are the estimated artist warping fields $F_{32} \in \mathbb{R}^{32 \times 32 \times 2}$ that, after bilinear upsampling to size $H \times W \times 2$, correspond to each artist caricature. We discuss this choice to select 32 $\times$ 32 as the warping field spatial size choice in the next section. To obtain these, we performed gradient descent optimization on the warping field for each $X_{toon}$ with $L1$ loss through the differentiable Warping Module to obtain the artist warping fields that correspond as closely as possible to each $X_{toon}$. To be precise, we solved the optimization 
\begin{equation}
    \operatorname*{argmin}_{F_{32}} ||X_{toon} - \textit{Warp}(X_{in}, \textit{Upsample}(F_{32}))||_1 \;.
    \label{eq:gt_warp}
\end{equation}

\subsection{Model Architecture}
AutoToon, our proposed method to tackle cartoon generation, is outlined in Figure \ref{fig:model}. The exaggeration network of AutoToon is comprised of two components: the Perceiver Network and Warping Module. The Perceiver Network is a truncated Squeeze-and-Excitation Network (SENet50)~\cite{hu2018senet} with weights pretrained on the VGGFace2 Dataset \cite{cao2018vggface2}, chosen due to its state-of-the-art facial recognition performance. In particular, we modify it by only keeping the original layers up to and including the second bottleneck block, followed by an adaptive average pooling layer with output size $32 \times 32 \times 2$. The purpose of truncating the network is to reduce network capacity and prevent overfitting to the small dataset. The Perceiver Network takes input image $X_{in}$ and outputs the warping field $\hat{F}_{32} \in \mathbb{R}^{32 \times 32 \times 2}$. $\hat{F}_{32}$ is then upsampled via bilinear upsampling to obtain $\hat{F}$, the per-pixel displacement. The Warping Module applies the warping field $\hat{F}$ to $X_{in}$ to obtain $\hat{X}_{toon}$. In inference, the warping field can also be multiplied by a scaling factor $\alpha$ to control the intensity of the warp, as shown in Figure \ref{fig:scaling}.

The choice to upsample a $32 \times 32$ warping field was motivated by two primary reasons. First, upsampling allows for an inherent smoothing of the warps, which intuitively creates smoother cartoons. Second, in keeping with powers of 2, a $64 \times 64$ warping field would have been too granular, and a $16 \times 16$ warping field was found to yield less exaggerated cartoons (see supplementary materials for details).

\subsection{Loss Functions}
We propose three loss functions to train AutoToon: the reconstruction loss, artist warping loss, and smoothness regularization loss.

The reconstruction loss $\mathcal{L}_{recon}$ penalizes the $L1$ distance between the artist cartoon $X_{toon}$ and the generated cartoon $\hat{X}_{toon}$. In addition to this supervision on the model output, we also supervise the warping fields themselves with the artist warping fields. The artist warping loss $\mathcal{L}_{warp}$ penalizes the $L1$ distance between the artist warping field $F_{32}$ obtained with \eqref{eq:gt_warp} and the estimated warping field $\hat{F}_{32}$. 

Finally, we use a cosine similarity regularization loss $\mathcal{L}_{reg}$ to encourage the warping field to be smooth and have fewer sudden changes in contour. This can be described as 
\begin{equation}
    \mathcal{L}_{reg} = \sum_{i,j \in \hat{\mathbf{F}} } \left( 2 - \frac{\langle \hat{\mathbf{F}}_{i,j-1}, \hat{\mathbf{F}}_{i,j} \rangle}{ \lVert \hat{\mathbf{F}}_{i,j-1} \rVert \lVert \mathbf{\hat{F}}_{i,j} \rVert }
    - \frac{\langle \hat{\mathbf{F}}_{i-1,j}, \hat{\mathbf{F}}_{i,j} \rangle}{ \lVert \hat{\mathbf{F}}_{i-1,j} \rVert \lVert \mathbf{\hat{F}}_{i,j} \rVert } \right),
\end{equation}
where $\langle \hat{\mathbf{F}}_{i,j-1}, \hat{\mathbf{F}}_{i,j} \rangle$ denotes the dot product of the upsampled warping field $\hat{\mathbf{F}}$ at pixel indices $i,j-1$ and $i,j$.

Thus, the loss function used to train our model is
\begin{equation}
    \mathcal{L}_{autotoon} = \lambda_1 \mathcal{L}_{recon} + \lambda_2 \mathcal{L}_{warp} + \lambda_3 \mathcal{L}_{reg}. \;
\end{equation}

\begin{figure*}
\begin{center}
\centering
\setlength{\tabcolsep}{0.2em}
\begin{tabular}{c|cccc|ccc}
    \includegraphics[width=0.117 \linewidth]{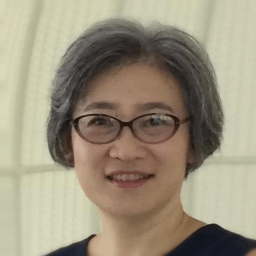} & \includegraphics[width=0.117 \linewidth]{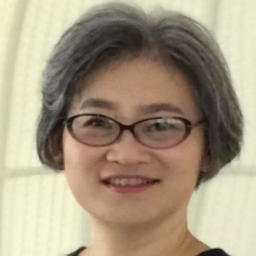} & \includegraphics[width=0.117 \linewidth]{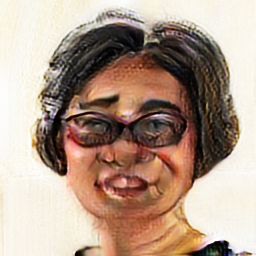} & \includegraphics[width=0.117 \linewidth]{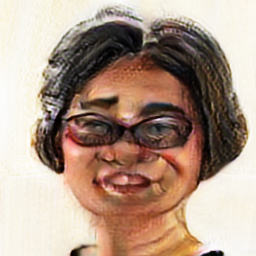} & \includegraphics[width=0.117 \linewidth]{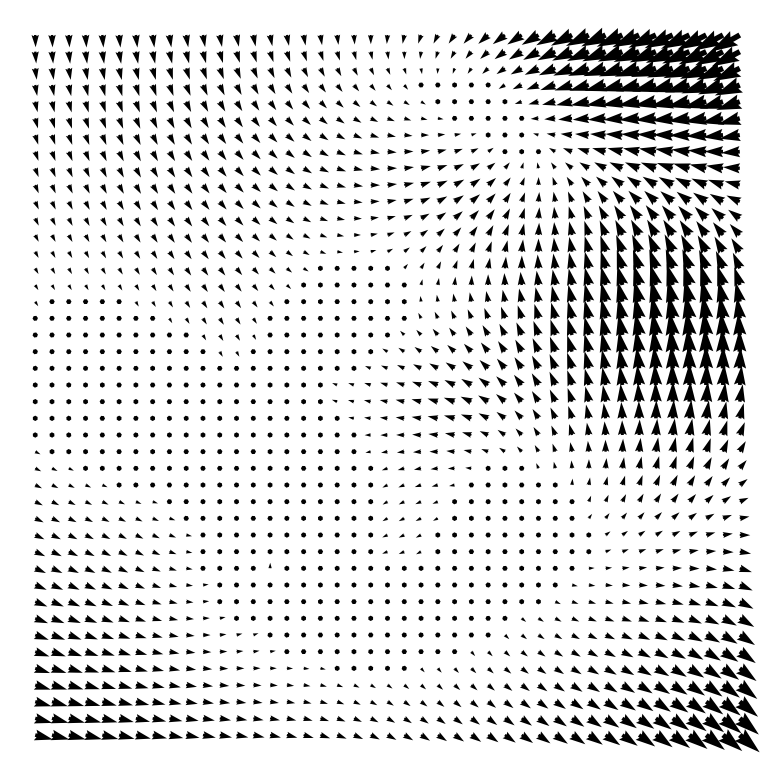} &\includegraphics[width=0.117 \linewidth]{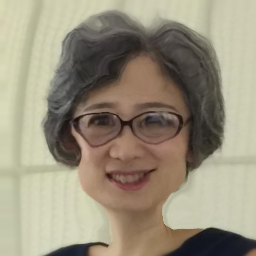} & \includegraphics[width=0.117 \linewidth]{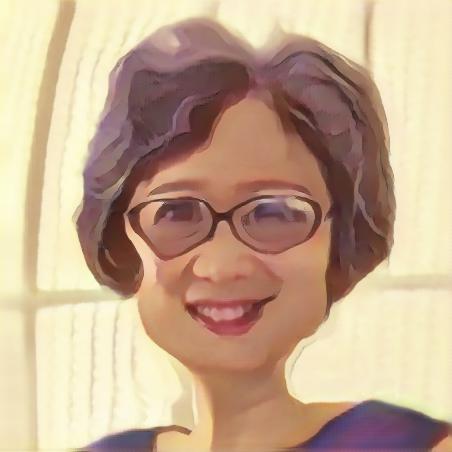} & \includegraphics[width=0.117 \linewidth]{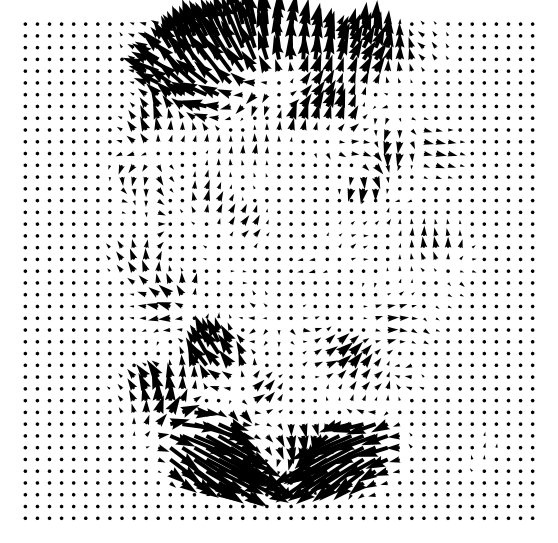} \\
    \includegraphics[width=0.117 \linewidth]{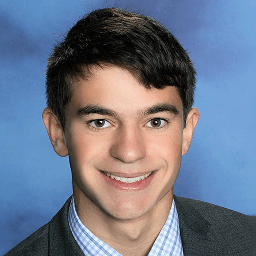} & \includegraphics[width=0.117 \linewidth]{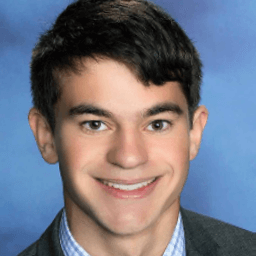} & \includegraphics[width=0.117 \linewidth]{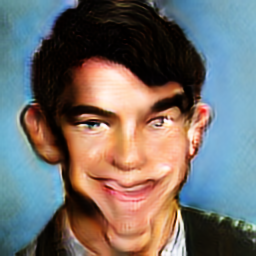} & \includegraphics[width=0.117 \linewidth]{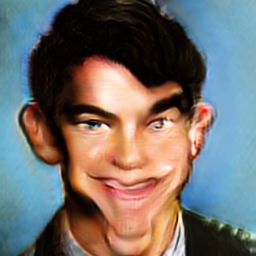} & \includegraphics[width=0.117 \linewidth]{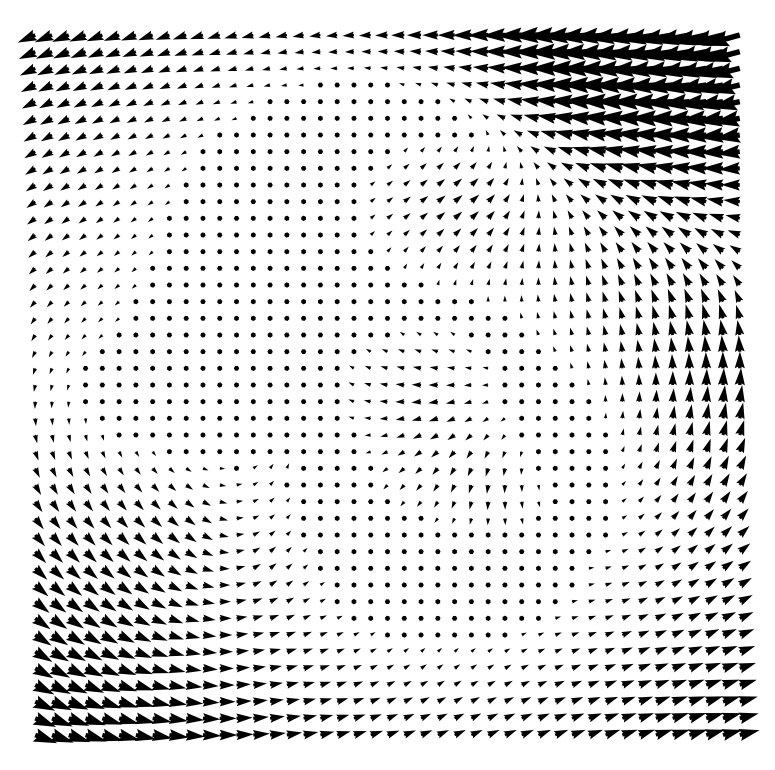} &\includegraphics[width=0.117 \linewidth]{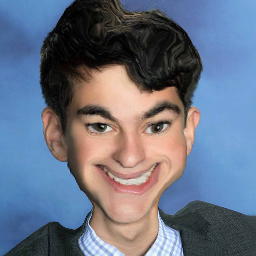} & \includegraphics[width=0.117 \linewidth]{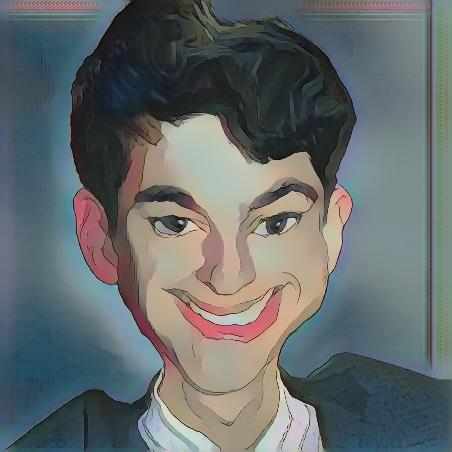} & \includegraphics[width=0.117 \linewidth]{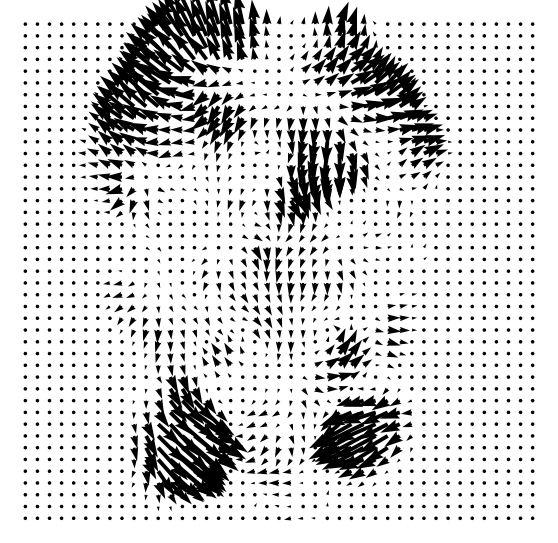} \\
    \includegraphics[width=0.117 \linewidth]{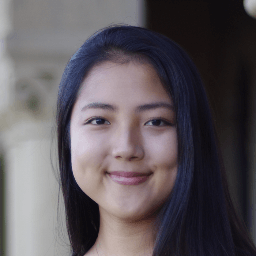} & \includegraphics[width=0.117 \linewidth]{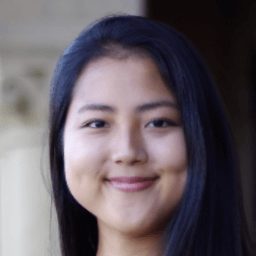} & \includegraphics[width=0.117 \linewidth]{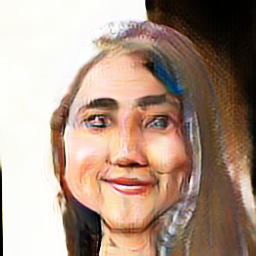} & \includegraphics[width=0.117 \linewidth]{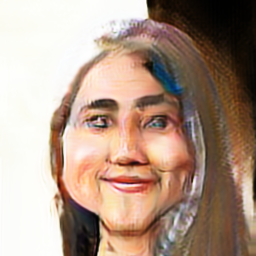} & \includegraphics[width=0.117 \linewidth]{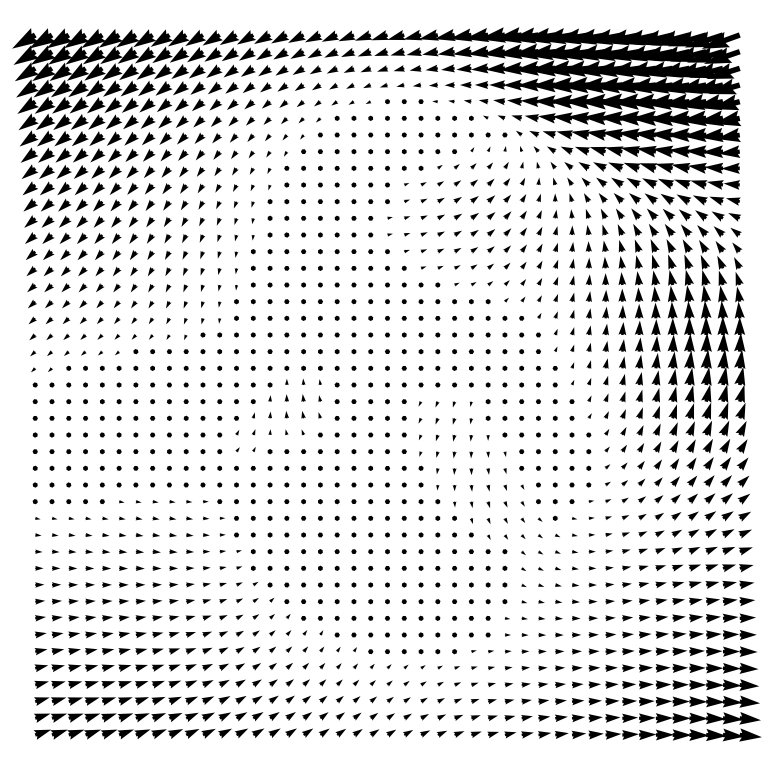} &\includegraphics[width=0.117 \linewidth]{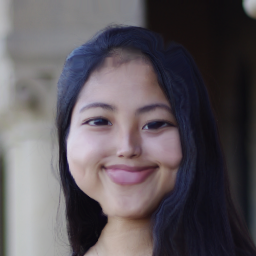} & \includegraphics[width=0.117 \linewidth]{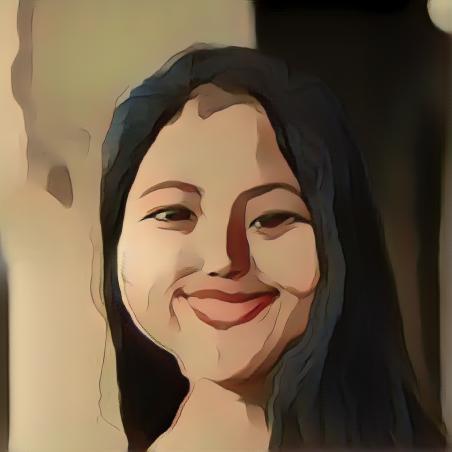} & \includegraphics[width=0.117 \linewidth]{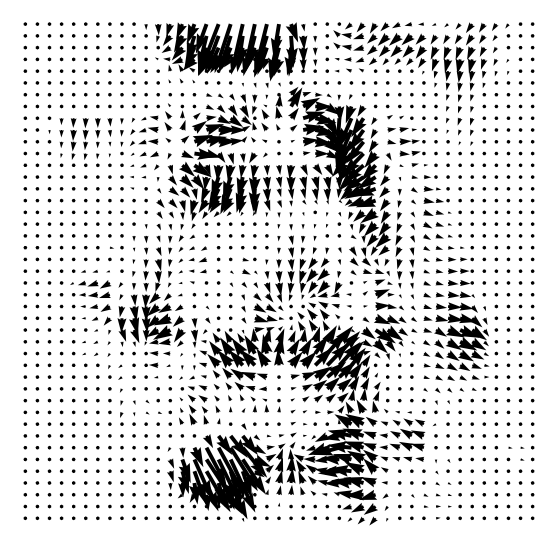} \\
    \includegraphics[width=0.117 \linewidth]{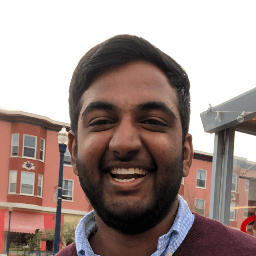} & \includegraphics[width=0.117 \linewidth]{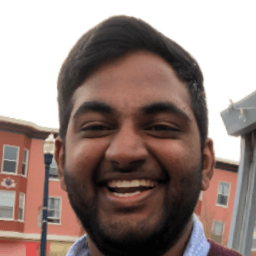} & \includegraphics[width=0.117 \linewidth]{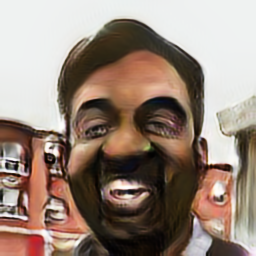} & \includegraphics[width=0.117 \linewidth]{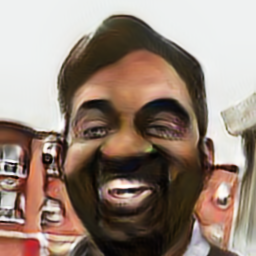} & \includegraphics[width=0.117 \linewidth]{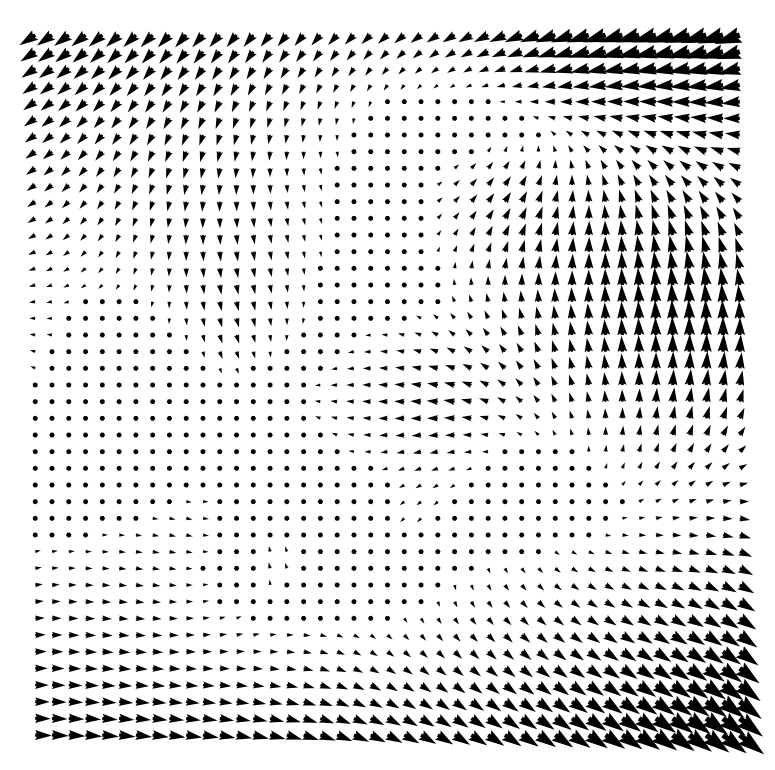} &\includegraphics[width=0.117 \linewidth]{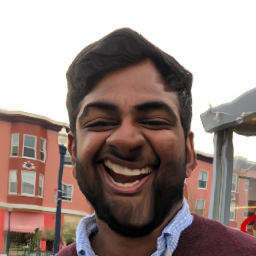} & \includegraphics[width=0.117 \linewidth]{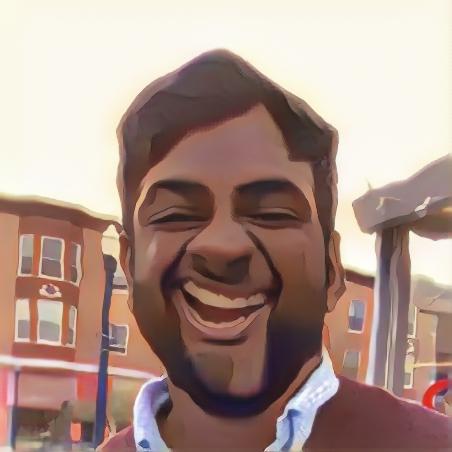} & \includegraphics[width=0.117 \linewidth]{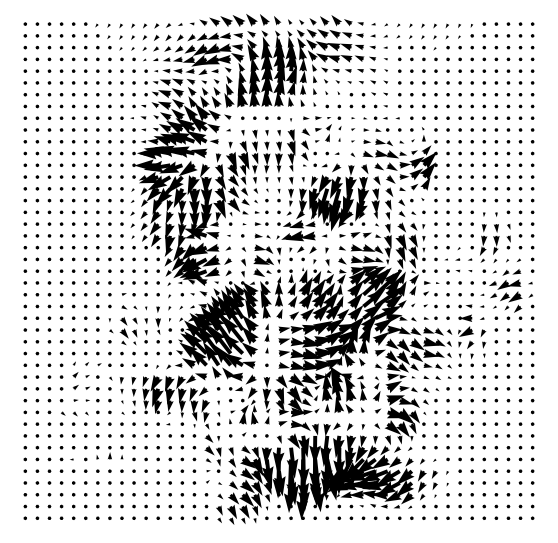} \\
    (a) & (b) & (c) & (d) & (e) & (f) & (g) & (h) \\
\end{tabular}
\end{center}
\vspace{-1.5em}
  \caption{Comparison of our method to WarpGAN \cite{shi2019warpgan} to visualize disentanglement of geometry from style. From left to right: input images (a) from our test set, (b) the result of passing the image through WarpGAN's warping module without performing stylization, (c) stylizing using WarpGAN's encoder and decoder, but without the warping module, (d) the final output of WarpGAN, and (e) the visualized WarpGAN warping fields. Then, we have exaggerated cartoons (f) generated by our model, our model's cartoons stylized with CartoonGAN \cite{ChenCartoonGAN2018} to create caricatures (g), and our visualized warping fields (h). See supplementary materials for more comparisons.}
\label{fig:disentangle}
\end{figure*}

\section{Experiments and Discussion}
\subsection{Training Details}
We use the Adam optimizer with $\beta_1 = 0.5$ and $\beta_2 = 0.999$, and with learning rate decay 0.95. With a batch size of 16, each minibatch consists of a randomly selected and aligned input-cartoon pair with the corresponding artist warp. Two types of online data augmentation are applied to the input images: random horizontal flips, as well as color jitter (brightness, contrast, and saturation jitter each uniformly sampled from the range $[0.9, 1.1]$ and hue jitter uniformly sampled from the range $[-0.05, 0.05]$ as specified by the PyTorch color jitter API). We empirically set $\lambda_1 = 1$, $\lambda_2 = 0.7$, and $\lambda_3 = 1e$-$6$. All experiments were conducted with PyTorch version 1.1 on Tesla V100 GPUs.

\subsection{Ablation Study}
\begin{figure}[!]
\begin{center}
\centering
\setlength{\tabcolsep}{0.1em}
\begin{tabular}{ccccc}
    \includegraphics[width=0.0916 \textwidth]{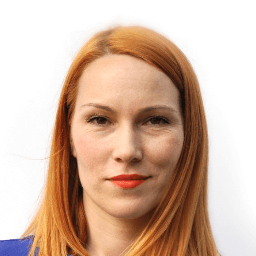} & \includegraphics[width=0.0916 \textwidth]{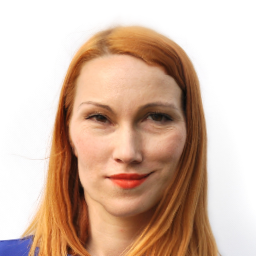} & \includegraphics[width=0.0916 \textwidth]{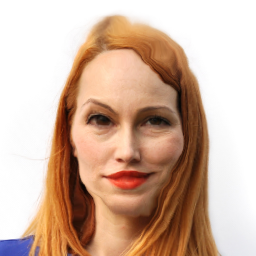} & \includegraphics[width=0.0916 \textwidth]{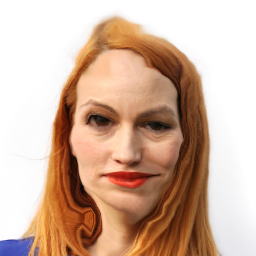} & \includegraphics[width=0.0916 \textwidth]{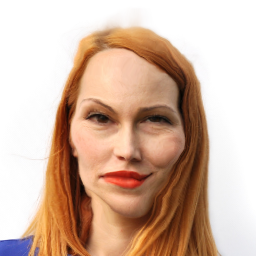} \\
    \includegraphics[width=0.0916 \textwidth]{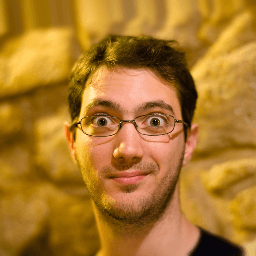} & \includegraphics[width=0.0916 \textwidth]{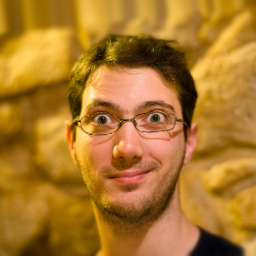} & \includegraphics[width=0.0916 \textwidth]{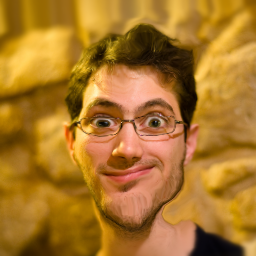} & \includegraphics[width=0.0916 \textwidth]{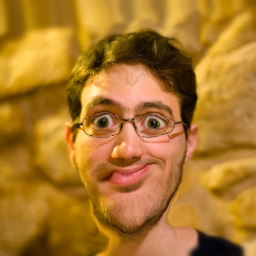} & \includegraphics[width=0.0916 \textwidth]{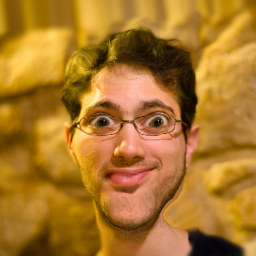} \\
    \includegraphics[width=0.0916 \textwidth]{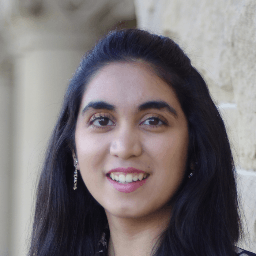} & \includegraphics[width=0.0916 \textwidth]{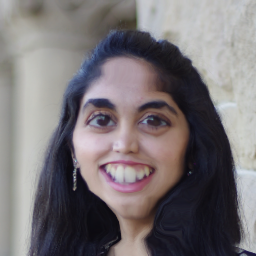} & \includegraphics[width=0.0916 \textwidth]{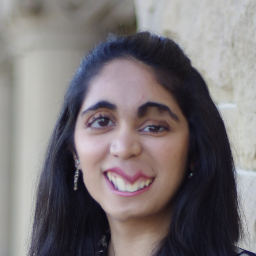} & \includegraphics[width=0.0916 \textwidth]{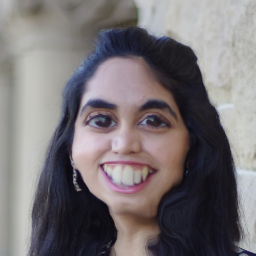} & \includegraphics[width=0.0916 \textwidth]{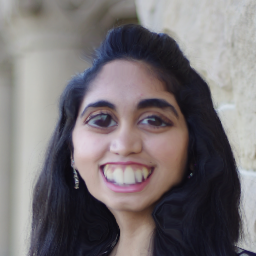} \\
    Input & No $\mathcal{L}_{warp}$ & No $\mathcal{L}_{recon}$ & No $\mathcal{L}_{reg}$ & With all \\
\end{tabular}
\end{center}
\vspace{-1.5em}
  \caption{Cartoons of model variations without each proposed loss on images from validation (first two) and test set (last). Photos 1 and 2 by \href{https://www.flickr.com/photos/pirati/35443782186/}{Pirátská strana} and \href{https://www.flickr.com/photos/fdevillamil/3533714277/}{Frédéric de Villamil}; modified.}
\label{fig:ablation}
\end{figure}

We train three additional variations of our model to analyze the contribution of each loss function to the system performance, as shown in Figure \ref{fig:ablation}. Without the artist warp loss, the warps are much weaker and constrained to detailed features, and they do not dramatically alter the face shape. Without the reconstruction loss, the warps are larger in scope, but twist the face dramatically to the point where it unnaturally distorts the face. Without the proposed cosine similarity regularization loss, the warping field is less smooth and introduces some implausible asymmetries, artifacts, and inconsistencies in the facial warping.

\begin{table}[!]
    \centering
    \setlength{\tabcolsep}{0.35em}
    \begin{tabular}{lcc}
        \toprule
        \textbf{Score} & \textbf{WarpGAN \cite{shi2019warpgan}} & \textbf{AutoToon} \\
        \midrule
        \textbf{Exaggeration} & 3.2 & \textbf{4.5} (p $<0.01$) \\
        \textbf{User Preference} & 30.1\% & \textbf{69.9\%} (p $<0.0001$) \\
        \bottomrule
    \end{tabular}
    \caption{Results (averages) of user studies for artists and casual observers. Artists rated images from 1 (worst) to 10 (best). Casual observers chose the image with more convincing exaggeration; the proportion of user selections for each model are shown here.}
    \label{tab:user_study}
\end{table}

\subsection{Warping Quality User Study}
We conducted two user studies to assess the quality of the warps learned by AutoToon. Since our contribution is purely the warping component of the caricature generation framework, we evaluated the quality of our warps against the performance of the warping module in the state-of-the-art, WarpGAN \cite{shi2019warpgan}.
For each of 24 images, we asked 14 trained artists to provide ratings of cartoons generated by each network from 1 (worst) to 10 (best) for exaggeration quality, or the exaggeration of the subject's most prominent features. We also asked 37 casual observers to select between AutoToon cartoons and WarpGAN cartoons for the more ``visually convincing'' cartoon for the subject. To ensure earnest responses, the participants were in a controlled setting and attentive to the task rather than randomly crowdsourced. These results are shown in Table \ref{tab:user_study}.

AutoToon consistently performs higher for both casual observers and artists (p $< 0.0001$ from 1-sample proportion test, p $< 0.01$ from 2-sample t-test respectively), making it a strong warping module for cartoon generation. We hypothesize that $\sim30\%$ of users preferred WarpGAN cartoons because these images are often warped so weakly that they nearly exactly match the original image (see Figure \ref{fig:disentangle}), creating such a stark contrast to AutoToon's that users perceived AutoToon's as distorted. The artists also provided feedback that they would have liked to see even more symmetry and less distortion in AutoToon warps, but that they preferred this to WarpGAN warps that did not alter specific facial features and only mildly stretched the image. We leave these improvements to future work.

\subsection{Disentanglement of Geometry and Style}

Disentangling warping and stylization is valuable for providing greater flexibility in combining warped images with different styles, as well as potential uses where only pure warping is desired to create photorealistic deformation. It also encourages preservation of details, as we will discuss shortly. A strong warping module is thus an important contribution to a complete caricature pipeline. AutoToon only performs geometric warping, so its output is photorealistic and can be separately stylized by any stylization method.

To evaluate AutoToon's warping quality, we can compare it to the warping module of WarpGAN \cite{shi2019warpgan} by evaluating the extent of disentanglement. In Figure \ref{fig:disentangle}, we examine WarpGAN's output image with only warping from its warping module, only stylization, both warping and stylization, and the corresponding warping field. We compare the warping-only cartoon to the output cartoon of AutoToon, the result of applying stylization to this output to create the final caricature, and the warping field learned by AutoToon.

We find that WarpGAN's cartoon images (b) do not significantly deviate from the input images (a), only providing relatively coarse and somewhat weak warping. The geometric differences between the stylized images (c) and the final caricatures (d) are also minimal. We can confirm that the warps are not that strong and do not provide a clear signal of distinguishing facial characteristics between identities by looking at the warping fields (e), which have very general shapes. Thus, WarpGAN's stylization network carries the majority of the geometric contribution to the final caricature in looking at the difference between the inputs (a) and stylized images (c).

In contrast, applying stylization (g) to AutoToon's outputs does not significantly alter the geometry of the cartoons (f), and geometric differences between (a) and the cartoons (f) are large, so the vast majority of the geometric contribution to the final caricature comes from AutoToon. Note also the strength and specificity of the warps learned by AutoToon in (h). Not only are the warps larger in magnitude and localized around facial features, but they are also clearly different for each identity on the level of facial features.

\begin{figure}[!]
\begin{center}
\centering
\setlength{\tabcolsep}{0.1em}
\begin{tabular}{ccccc}
    \footnotesize
    \shortstack{Input Photo \\ \quad \\ \quad \\ \quad \\ \quad \\ \quad} & \includegraphics[width=0.182 \linewidth]{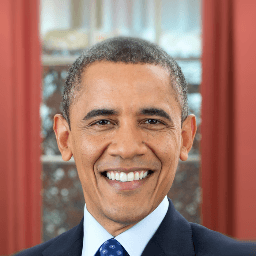} & \includegraphics[width=0.182 \linewidth]{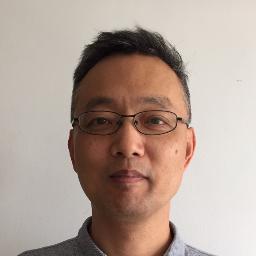} & \includegraphics[width=0.182 \linewidth]{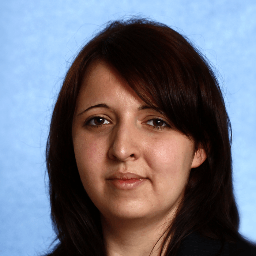} & \includegraphics[width=0.182 \linewidth]{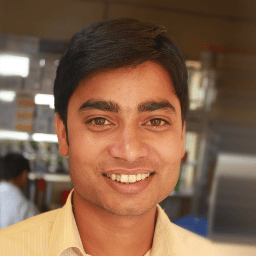} \\
    
    \footnotesize
    \shortstack{AutoToon \\ \quad \\ \quad \\ \quad \\ \quad \\ \quad \\ \quad} & \includegraphics[width=0.182 \linewidth]{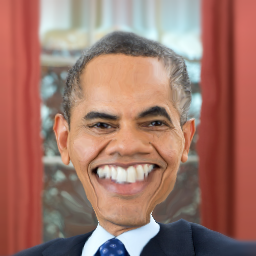} & \includegraphics[width=0.182 \linewidth]{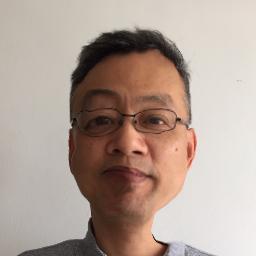} & \includegraphics[width=0.182 \linewidth]{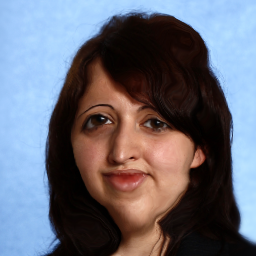} & \includegraphics[width=0.182 \linewidth]{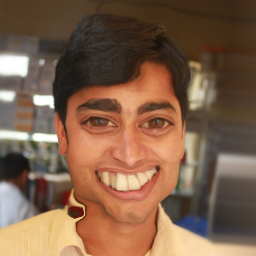} \\
    
    \footnotesize
    \shortstack{AutoToon + \\ CartoonGAN \cite{ChenCartoonGAN2018} \\ \quad \\ \quad \\ \quad \\ \quad} & \includegraphics[width=0.182 \linewidth]{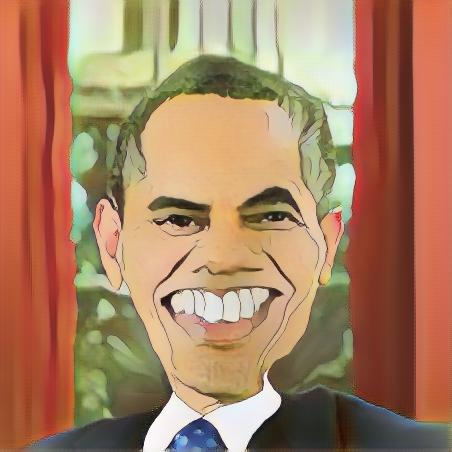} & \includegraphics[width=0.182 \linewidth]{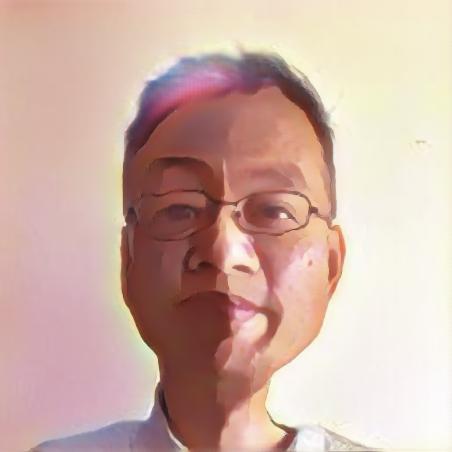} & \includegraphics[width=0.182 \linewidth]{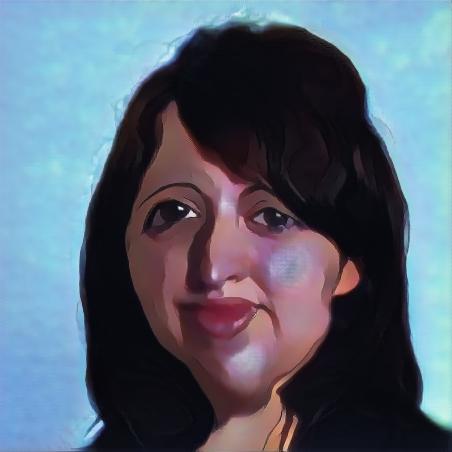} & \includegraphics[width=0.182 \linewidth]{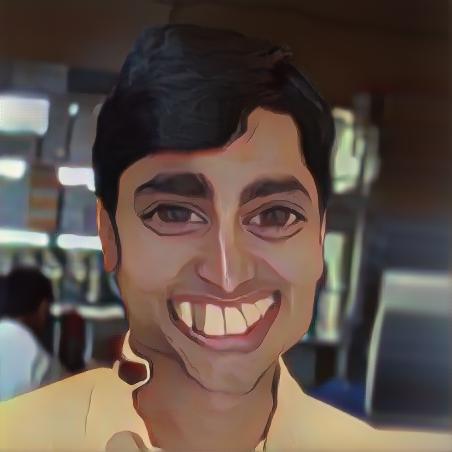} \\
    
    \footnotesize
    \shortstack{WarpGAN \cite{shi2019warpgan} \\ \quad \\ \quad \\ \quad \\ \quad \\ \quad \\ \quad} & \includegraphics[width=0.182 \linewidth]{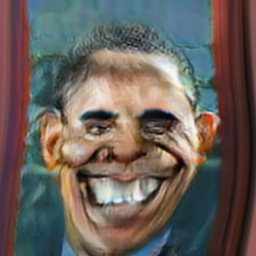} & \includegraphics[width=0.182 \linewidth]{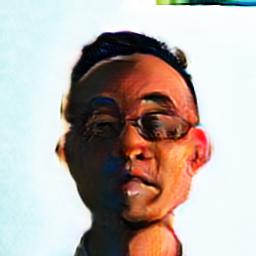} & \includegraphics[width=0.182 \linewidth]{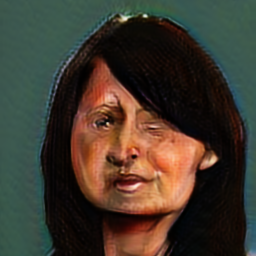} & \includegraphics[width=0.182 \linewidth]{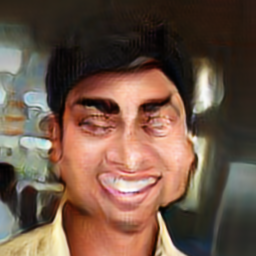} \\
\end{tabular}
\end{center}
\vspace{-1.5em}
  \caption{Comparison of AutoToon to WarpGAN \cite{shi2019warpgan} w.r.t. facial detail preservation (first two test, last two validation images). Photos 3 and 4 by \href{https://www.flickr.com/photos/robby_schulze/8791363560/}{Robby Schulze} and \href{https://www.flickr.com/photos/nyayahealth/7652902938/}{Possible}; modified.}
\label{fig:preserve_detail}
\end{figure}

\subsection{Preservation of Facial Detail}
Caricatures need not sacrifice visual quality of the input image when exaggerating salient facial characteristics. However, due to the incomplete disentanglement of geometry and style in WarpGAN, there exists an inherent tradeoff between stylization and facial detail preservation. As shown in Figure \ref{fig:preserve_detail}, WarpGAN's style is inseparable from its warping, creating inconsistencies or sacrificing details of the eyes, lowering the caricature quality. On the other hand, AutoToon exaggerates yet still preserves the overall quality and consistency of facial features in a way that is faithful to the original image, especially with respect to details such as the eyes, ears, and teeth. This is especially noteworthy because of the difficulty of convincingly preserving facial detail in a photorealistic image due to the lack of stylization that could potentially compensate for any warping artifacts.

\begin{figure}[!]
\begin{center}
\centering
\setlength{\tabcolsep}{0.1em}
\begin{tabular}{ccccc}
    \includegraphics[width=0.1155 \textwidth]{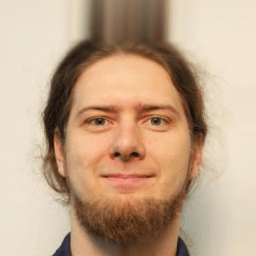} & \includegraphics[width=0.1155 \textwidth]{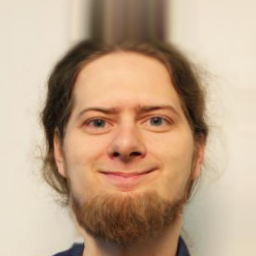} &
    \includegraphics[width=0.1155 \textwidth]{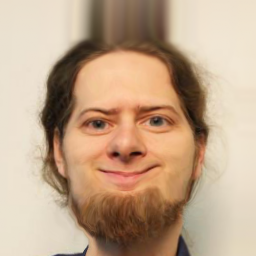} &
    \includegraphics[width=0.1155 \textwidth]{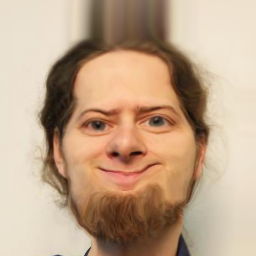} \\
    \includegraphics[width=0.1155 \textwidth]{originals/Archana.png} & \includegraphics[width=0.1155 \textwidth]{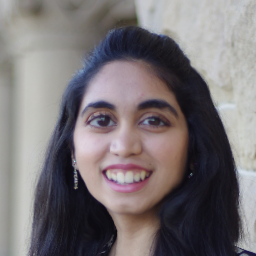} &
    \includegraphics[width=0.1155 \textwidth]{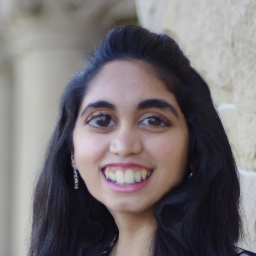} &
    \includegraphics[width=0.1155 \textwidth]{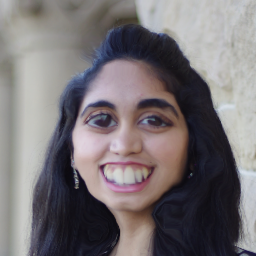} \\
    \includegraphics[width=0.1155 \textwidth]{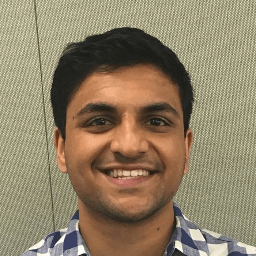} & \includegraphics[width=0.1155 \textwidth]{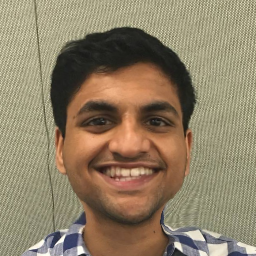} &
    \includegraphics[width=0.1155 \textwidth]{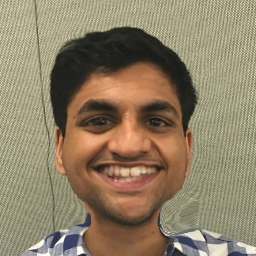} &
    \includegraphics[width=0.1155 \textwidth]{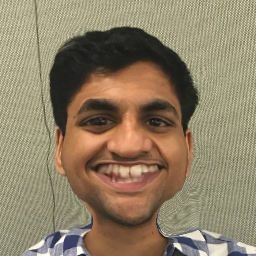} \\
    Input & $\alpha=1$ & $\alpha=1.5$ & $\alpha=2$ \\
\end{tabular}
\end{center}
\vspace{-1.5em}
  \caption{Result of scaling the warping field of various examples from the test set with scaling factor $\alpha$.}
\label{fig:scaling}
\end{figure}

It is also interesting to note that while AutoToon preserves facial plausibility, it is also in ``toon'' with facial asymmetries. For example, in Figure \ref{fig:scaling}, the second subject's left eye (from their perspective) is slightly smaller than their right; with increases in the scaling factor $\alpha$, this asymmetry is amplified. We also see similar amplifications for the crooked smile of subject 4 in Figure \ref{fig:disentangle} and the smirks of subjects 1 and 4 in Figure \ref{fig:learned_features}. This sort of exaggeration of asymmetry is crucial for creating caricatures because they often mark distinguishing features in individuals' faces.

\subsection{AutoToon Warp Transfer}
\begin{figure}[!]
\begin{center}
\centering
\setlength{\tabcolsep}{0.1em}
\begin{tabular}{cccc}
    \includegraphics[width=0.1155 \textwidth]{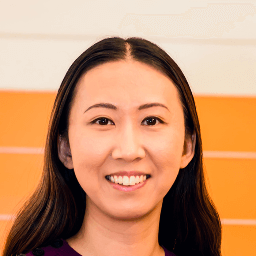} & \includegraphics[width=0.1155 \textwidth]{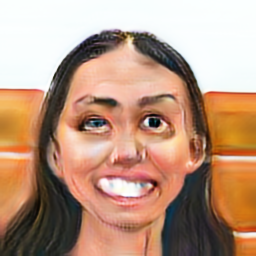} & \includegraphics[width=0.1155 \textwidth]{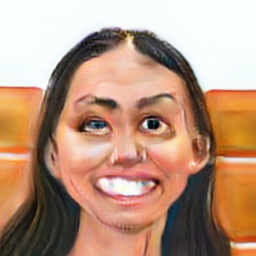} & \includegraphics[width=0.1155 \textwidth]{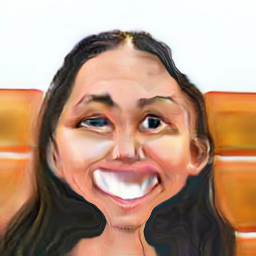} \\
    \includegraphics[width=0.1155 \textwidth]{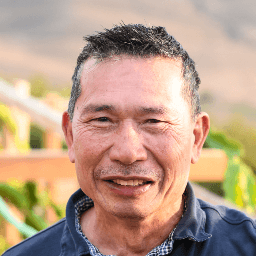} & \includegraphics[width=0.1155 \textwidth]{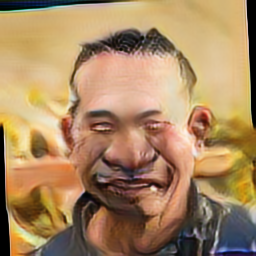} & \includegraphics[width=0.1155 \textwidth]{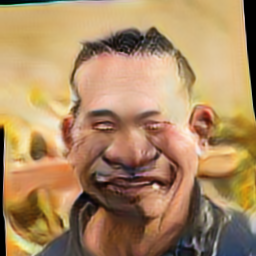} & \includegraphics[width=0.1155 \textwidth]{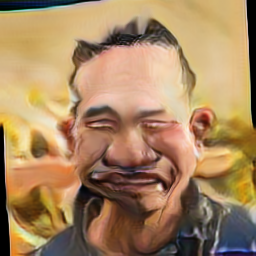} \\
    \includegraphics[width=0.1155 \textwidth]{originals/Huifang.png} & \includegraphics[width=0.1155 \textwidth]{warpgan_style/Huifang.png} & \includegraphics[width=0.1155 \textwidth]{caricatures/warpgan/Huifang.png} & \includegraphics[width=0.1155 \textwidth]{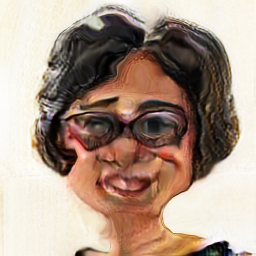} \\
    (a) & (b) & (c) & (d) \\
\end{tabular}
\end{center}
\vspace{-1.5em}
  \caption{Input images (a) from test (first, third) and validation set (second), unwarped, stylized images generated by WarpGAN \cite{shi2019warpgan} (b), warped caricatures generated by WarpGAN (c), and (d), result of applying the warping fields generated by AutoToon to (b).}
\label{fig:warpgan_ourwarp}
\end{figure}

To illustrate the efficacy of AutoToon warps, we show in Figure \ref{fig:warpgan_ourwarp} the effect of applying AutoToon warps to stylized WarpGAN test images, in comparison to the end-to-end WarpGAN caricatures. The resulting images have stronger warps that enhance the prominent features of the subjects.

The warping quality of AutoToon warps can also be observed through manipulating the scaling factor $\alpha$, which scales the magnitude of the warping field used to generate the cartoons as shown in Figure \ref{fig:scaling}. Larger scale factors create more intense exaggerations, but still remain plausible and maintain the overall warping quality.

\subsection{Facial Feature-Specific Warping}
\begin{figure*}[!t]
\begin{center}
\centering
\setlength{\tabcolsep}{0.15em}
\begin{tabular}{ccccccc}
    & Big eyes & Thin neck & Thin eyes & Chiseled face & Big nose, hair & Big eyes \\
    
    & Long chin & Bulging cheeks & Wide mouth & Thick lips & Wide smile & Round face \\

    \shortstack{Input Photo \\ \quad \\ \quad \\ \quad \\ \quad \\ \quad \\ \quad \\ \quad \\ \quad \\ \quad \\ \quad} & \includegraphics[width=0.13 \linewidth]{originals/00005_flip.png} &
    \includegraphics[width=0.13 \linewidth]{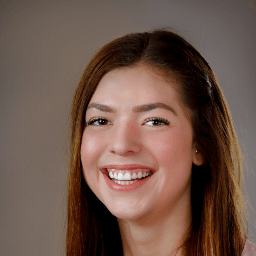} & \includegraphics[width=0.13 \linewidth]{originals/00000.png} & \includegraphics[width=0.13 \linewidth]{originals/00006.png} & \includegraphics[width=0.13 \linewidth]{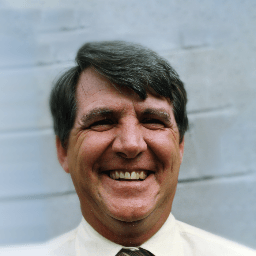} 
    & \includegraphics[width=0.13 \linewidth]{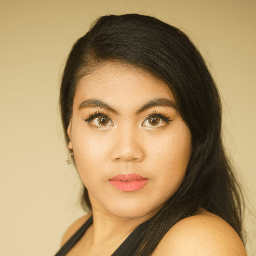} \\
    
    \shortstack{AutoToon \\ \quad \\ \quad \\ \quad \\ \quad \\ \quad \\ \quad \\ \quad \\ \quad \\ \quad \\ \quad} & \includegraphics[width=0.13 \linewidth]{cartoons/autotoon/00005_flip.png} & \includegraphics[width=0.13 \linewidth]{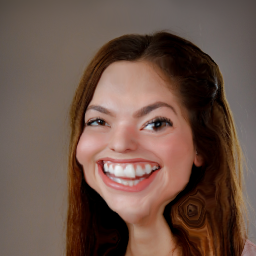} & \includegraphics[width=0.13 \linewidth]{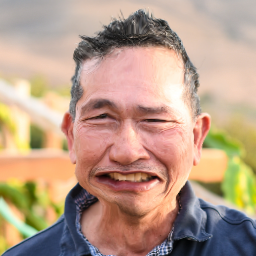} & \includegraphics[width=0.13 \linewidth]{cartoons/autotoon/00006.png} & \includegraphics[width=0.13 \linewidth]{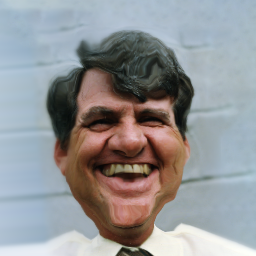} & \includegraphics[width=0.13 \linewidth]{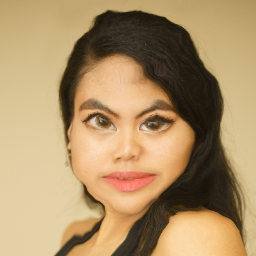} \\
    
    \shortstack{Artist \\ \quad \\ \quad \\ \quad \\ \quad \\ \quad \\ \quad \\ \quad \\ \quad \\ \quad \\ \quad} & \includegraphics[width=0.13 \linewidth]{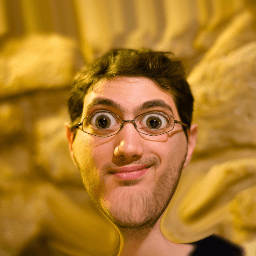} & \includegraphics[width=0.13 \linewidth]{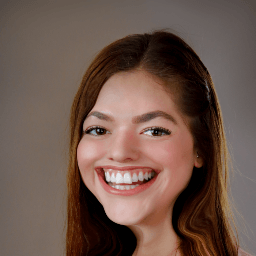} & \includegraphics[width=0.13 \linewidth]{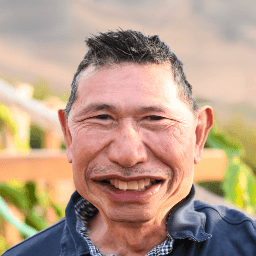} & \includegraphics[width=0.13 \linewidth]{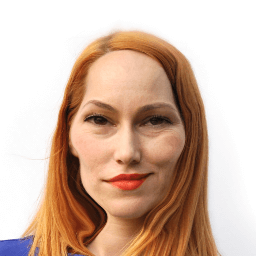} & \includegraphics[width=0.13 \linewidth]{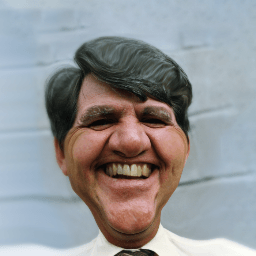} & \includegraphics[width=0.13 \linewidth]{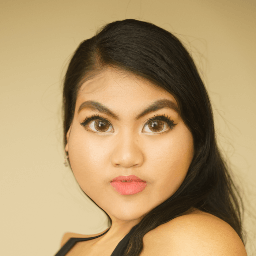} \\
\end{tabular}
\end{center}
\vspace{-1.5em}
  \caption{Examples of different detailed, face feature-specific exaggerations on the validation set learned by AutoToon as compared to artist cartoons. Shown are the input images, cartoons generated by our model, and the corresponding artist cartoons for the same subject. See supplementary materials for more results. Photos 2, 5, 6 by \href{https://www.flickr.com/photos/jayburg/31998771523/}{Jacob Seedenburg}, \href{https://flic.kr/p/25n98a6}{Community Archives}, and \href{https://www.flickr.com/photos/rankingfuuta/40937230841/}{Aaron Stidwell}; modified.}
\label{fig:learned_features}
\end{figure*}

Despite the small dataset size, AutoToon has learned a diverse range of warping styles, and in particular, specific facial feature-level exaggerations that are distinct for different individuals. Examples of different learned facial feature warps are shown in Figure \ref{fig:learned_features}. Many other examples exist, including the curved smile of the second individual in Figure \ref{fig:disentangle}. In contrast to previous work that utilizes sparse warping, this more granular level of amplification helps to bring out more nuanced features of an individual's face beyond a rough exaggeration of face shape.

\subsection{Face Pose Generalization}
\begin{figure}[H]
\begin{center}
\centering
\setlength{\tabcolsep}{0.1em}
\begin{tabular}{cccc}
    \scalebox{-1}[1]{\includegraphics[width=0.1155 \textwidth]{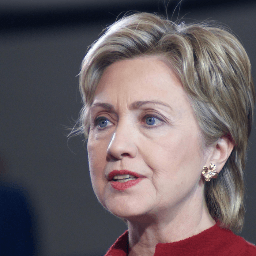}} & \scalebox{-1}[1]{\includegraphics[width=0.1155 \textwidth]{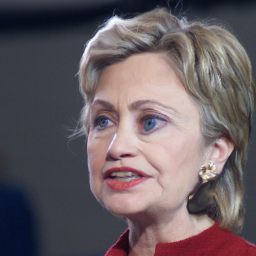}} &
    \includegraphics[width=0.1155 \textwidth]{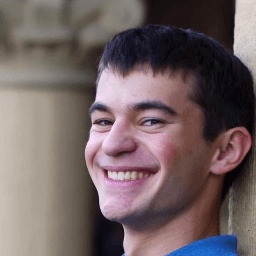} & \includegraphics[width=0.1155 \textwidth]{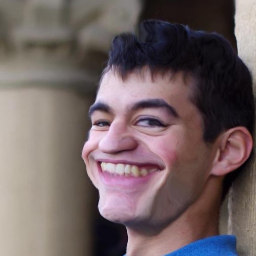} \\
    Input & Cartoon & Input & Cartoon \\
\end{tabular}
\end{center}
\vspace{-1.5em}
  \caption{Model generalization to non-frontal test set images.}
\label{fig:sideviews}
\end{figure}

Though only trained and validated on frontal-facing images, AutoToon performs relatively robustly on images in the test set with subjects that deviate from the frontal pose, shown in Figure \ref{fig:sideviews}. This suggests that the Perceiver Network has successfully captured face features that are robust to changes in angle and position.

\vspace{-0.5em}

\subsection{Visualization of Network Attention}

\begin{figure}[!]
\begin{center}
\centering
\setlength{\tabcolsep}{0.1em}
\begin{tabular}{cccc}
    \includegraphics[width=0.115\textwidth]{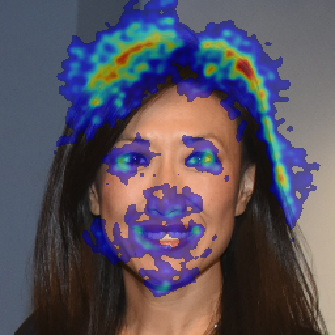} &
    \includegraphics[width=0.115\textwidth]{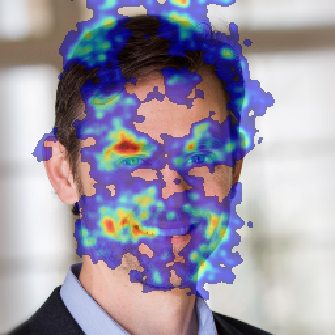} &
    \includegraphics[width=0.115\textwidth]{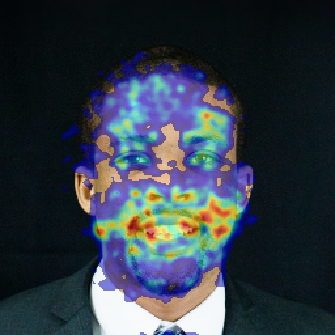} &
    \includegraphics[width=0.115\textwidth]{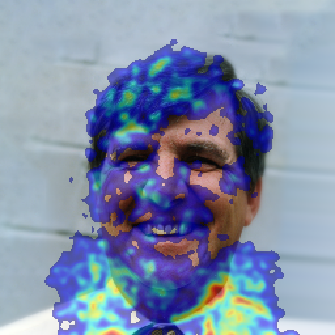} \\
    \includegraphics[width=0.115\textwidth]{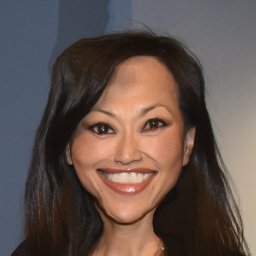} &
    \includegraphics[width=0.115\textwidth]{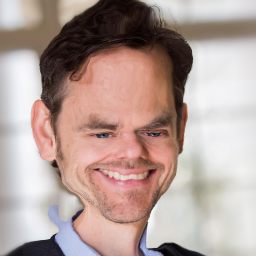} &
    \includegraphics[width=0.115\textwidth]{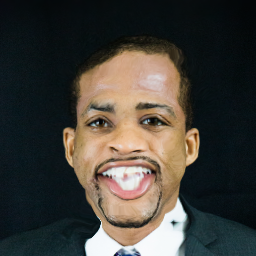} &
    \includegraphics[width=0.115\textwidth]{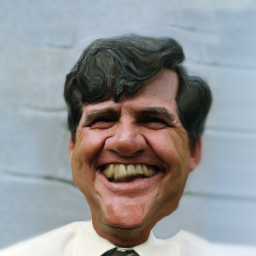} \\
    (a) & (b) & (c) & (d) \\
\end{tabular}
\end{center}
\vspace{-1.5em}
  \caption{First row: network attention with smoothed guided backpropagation~\cite{smilkov2017smoothgrad,springenberg2014striving} jet-overlaid on validation images (increasing from blue to red). Second row: generated cartoons. Our model focuses on specific features for each face, such as (a) hair and eyes, (b) eyes and smile dimples, (c) mouth, and (d) chin and neck. Photos 1 and 2 by \href{https://www.flickr.com/photos/mdgovpics/17879615876/}{Maryland GovPics} and \href{https://www.flickr.com/photos/silvery/40286723163/}{Si1very}; modified.}
\label{fig:smoothed_guided_backprop}
\end{figure}

In order to get a sense of the features used by our method to generate cartoons, we employ guided backpropagation~\cite{springenberg2014striving} that we couple with smoothgrad~\cite{smilkov2017smoothgrad} for a more stable analysis. We visualize the result of this analysis on 4 different images from our validation set in Figure~\ref{fig:smoothed_guided_backprop}.

\subsection{Limitations}
\begin{figure}[t]
\begin{center}
\centering
\setlength{\tabcolsep}{0.1em}
\begin{tabular}{cccc}
    \shortstack{(a) \\ \quad \\ \quad \\ \quad \\ \quad \\ \quad \\ \quad \\ \quad} & \includegraphics[width=0.12 \textwidth]{originals/00009.png} & \includegraphics[width=0.12 \textwidth]{cartoons/autotoon/00009.png} & \includegraphics[width=0.12 \textwidth]{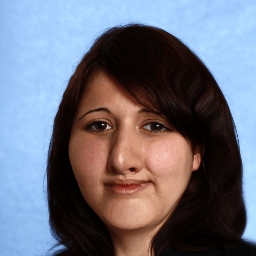} \\
    \shortstack{(b) \\ \quad \\ \quad \\ \quad \\ \quad \\ \quad \\ \quad \\ \quad} & \includegraphics[width=0.12 \textwidth]{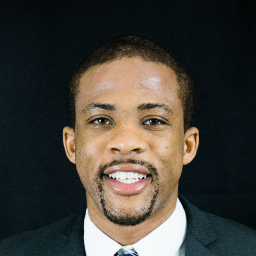} & \includegraphics[width=0.12 \textwidth]{cartoons/autotoon/00008_flip.png} & \includegraphics[width=0.12 \textwidth]{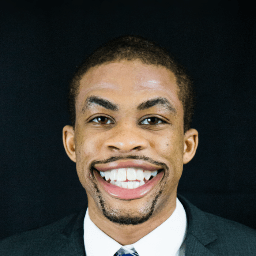} \\
    & Input & AutoToon & Artist \\
\end{tabular}
\end{center}
\vspace{-1.5em}
  \caption{Model limitations illustrated by examples from the validation set, consistent with artist comments from the user study. Photo 2 by \href{https://www.flickr.com/photos/vrcarabeo/10232772063/}{Vince Crabeo}; modified.}
\label{fig:limitations}
\end{figure}

Some limitations of AutoToon are illustrated in Figure \ref{fig:limitations}. Compared to the ground-truth image, the model incorrectly enlarges the eyes in (a), likely because bulging of eyes is very common in the dataset. The chin in (a) and mouth and eyebrows of (b) are not as successfully warped and introduce some distortion and warping artifacts.

\section{Conclusion}
In this paper, we present AutoToon, the first supervised deep learning method for cartoonization, or the warping step of facial caricature generation. Our warping method yields high-quality warps that outperform the state-of-the-art. Our model is also disentangled entirely from style, allowing it to be paired with any stylization network, including existing caricature generation models, to create diverse caricatures. Unlike previous work, it leverages the power of the SENet and differentiable warping module, and also learns directly from artist warping fields. In addition to creating convincing exaggerations that are subject- and facial feature-specific, it also preserves facial detail faithful to the original image and generalizes to non-frontal portrait images. We evaluated these caricatures qualitatively in comparison to prior art with respect to geometry and style disentanglement, facial detail preservation, and warping quality and feature-level specificity, and quantitatively showed through our user study and artist ratings that AutoToon outperforms state-of-the-art networks in geometric warping. Future directions of interest include further smoothing of the warping field to avoid pixel collision, identity preservation, and few-shot learning to adapt to different artist styles.

{\small
\bibliographystyle{ieee}
\bibliography{bibliography}
}

\end{document}